Spatial Transcriptomics Expression Prediction from Histopathology Based on Cross-Modal Mask Reconstruction and Contrastive Learning

Junzhuo Liu[1], Markus Eckstein[2, 3, 4, 5], Zhixiang Wang[6], Friedrich Feuerhake[7,8, *], Dorit Merhof [1,9] *

[1] Faculty of Informatics and Data Science, University of Regensburg, Regensburg, Germany

[2] Institute of Pathology, Universitätsklinikum Erlangen, Friedrich-Alexander-Universität Erlangen-Nürnberg (FAU), Erlangen, Germany

[3] Comprehensive Cancer Center Erlangen-EMN (CCC ER-EMN), Erlangen, Germany

[4] Comprehensive Cancer Center Alliance WERA (CCC WERA), Erlangen, Germany

[5] Bavarian Cancer Research Center (BZKF), Erlangen, Germany

[6] Department of Ultrasound, Beijing Friendship Hospital, Capital Medical University, Beijing, China

[7] Institute for Pathology, Hanover Medical School, Hannover, Germany

[8] Institute for Neuropathology, University Clinic Freiburg, Freiburg i.Br., Germany

[9] Fraunhofer Institute for Digital Medicine MEVIS, Bremen, Germany

**Abstract**

Spatial transcriptomics is a technology that captures gene expression levels at different spatial locations, widely used in tumor microenvironment analysis and molecular profiling of histopathology, providing valuable insights into resolving gene expression and clinical diagnosis of cancer. Due to the high cost of data acquisition, large-scale spatial transcriptomics data remain challenging to obtain. In this study, we develop a contrastive learning-based deep learning method to predict spatially resolved gene expression from whole-slide images. Evaluation across six different disease datasets demonstrates that, compared to existing studies, our method improves Pearson Correlation Coefficient (PCC) in the prediction of highly expressed genes, highly variable genes, and marker genes by 6.27%, 6.11%, and 11.26% respectively. Further analysis indicates that our method preserves gene-gene correlations and applies to datasets with limited samples. Additionally, our method exhibits potential in cancer tissue localization based on biomarker expression.

**Introduction**

Cancer is one of the leading causes of mortality worldwide, with approximately one in five men or women developing cancer during their lifetime[1]. In 2020, nearly 20 million new cancer cases and approximately 10 million cancer-related deaths were reported globally. Demographic projections indicate a rising trend in these numbers, continuously imposing significant burdens on public health and socioeconomic systems[2]. In clinical practice, cancer is usually diagnosed based on microscopic analysis of tissue obtained by surgical resection or biopsy. The current standard for tumor classification defined by the WHO is based on histopathology and increasingly on additional molecular markers such as DNA sequencing, fluorescence in-situ hybridization (FISH), and DNA methylation analysis, defining histogenesis (cellular origin) and degree of malignancy (histopathological grade)[3]. In addition, the tumor size and its spreading into surrounding tissues (locally invasive growth) and distant organs (metastatic spread) are classified by the IUCC in the TNM classification system[4]. Cancer is a complex disease[5] caused by dysregulation of multiple cellular functions referred to as "hallmarks of cancer"[6]. In specific types of cancer, histopathology provides an intuitive visualization of tumor cell morphology[7-9], tissue architecture[10] and differentiation status[11,12], offering critical clinical insights for cancer diagnosis[13,14], tumor grading[15] and prognostic assessment[16,17]. The advent of spatial transcriptomics (ST) has provided a molecular-level complement to histopathology-based diagnostics. ST

These authors jointly supervised this work: Friedrich Feuerhake, Dorit Merhof. E-mail: Feuerhake.Friedrich@mh-hannover.de; Dorit.Merhof@informatik.uni-regensburg.de

not only reveals spatial transcriptional patterns and regulatory mechanisms within tissues[18], but also captures the tumor microenvironment and local tissue characteristics[19,20], offering novel insights for precision oncology.

The acquisition of spatial transcriptomics data typically involves tissue sample sectioning, tissue dissociation, and the integration of barcodes and index[21]. The high cost of specialized equipment and the complexity of the preparation process pose significant challenges for generating large-scale spatial transcriptomic data from whole-slide images (WSIs) in clinical practice[22]. Therefore, Deep Learning models that predict spatial transcriptomic expression directly from histopathology images offer a promising solution to this limitation. STNet[23] was one of the pioneering methods for modeling the correlation between tissue morphology and spatial transcriptional expression using convolutional neural networks. It employed a convolutional architecture to hierarchically extract deep visual features from histopathological images and map these features to spatial gene expression in an end-to-end manner. HisToGene[24] enhanced spatial dependency modeling of spatial transcriptomic measurement spots by leveraging a Transformer-based framework. Similarly, EGGN[25] utilized graph neural networks (GNNs) to model variations in gene expression levels across different spatial windows, dynamically improving gene expression prediction.

Despite the promising potential of these methods in predicting spatially resolved gene expression by transcriptomics from histopathological images, several limitations and challenges remain: 1. Most end-to-end models inevitably rely on fixed gene labels[26,27]. These methods select target genes based on their average expression levels or variability across the entire dataset. However, even within the same dataset, spatial transcriptomic gene expression levels and variability can vary significantly among different samples. As a result, genes that exhibit high expression or variability only in a subset of samples may not be included in the fixed prediction targets, thereby reducing the flexibility of gene expression prediction. Recently, BLEEP[28] was introduced to improve prediction flexibility based on contrastive learning. BLEEP models the relationship between histopathological morphological features and spatial transcriptomic expression through contrastive learning by computing the similarity between cross-modal data in the feature space to generate gene expression predictions from histopathological images. However, previous studies suggest that multimodal features often reside in distinct representational spaces after being encoded by multimodal encoders, making it challenging for the encoder to capture their interactions effectively[29]. 2. Current methods have not proposed effective strategies for integrating histopathological images with spatial transcriptomic data. Existing end-to-end models treat spatial transcriptomic data merely as labels while using histopathology images as input. BLEEP[28] implements cross-modal contrastive learning based on an individual discrimination task[30], but does not introduce further feature fusion strategies. 3. Most existing approaches are validated on only a single disease dataset and do not adopt cross-validation[23,24]. Our research indicates that, even within the same dataset cohort, different samples exhibit varying degrees of prediction difficulty. Validation based on only a subset of samples often results in unreliable performance evaluation.

To address these limitations and to model the relationship between histopathological morphological features and spatial transcriptomic expression levels for gene expression prediction from images, we propose a novel contrastive learning framework based on cross-modal mask reconstruction, named CMRCNet. The overall architecture of CMRCNet is shown in Figure 1. CMRCNet comprises three stages: contrastive learning, cross-modal reconstruction, and similarity-based inference. In the contrastive learning stage, an image encoder and a gene encoder extract features from paired histopathological image patches and spatial transcriptomic expression data, respectively, constructing embeddings for the similarity matrix used in contrastive learning. The contrastive loss function serves as an initial mechanism to align the feature representations of the two modalities in the spatial space. To bridge the gap between different modalities in the feature space and address the challenge that encoders struggle to capture cross-modal interactions, we introduce cross-modal reconstruction as a pretext task. This task represents both features within a unified space, facilitating effective interaction between histopathological morphology and gene expression levels. Specifically, in the cross-modal reconstruction stage, randomly masked image features are progressively reconstructed under the guidance of gene expression features. We propose a cross-modal reconstruction module based on a Transformer architecture to ensure consistency in image feature encoding. Within this module, we design a cross-attention mechanism with residual mapping to restore image features from gene encodings. This attention mechanism not only prevents the reconstruction process from overly relying on a single modality, but also ensures that unmasked morphological features and fused information at each stage contribute to the reconstruction. We evaluate our method on six datasets across different diseases, predicting highly expressed genes, highly variable genes, and marker genes, demonstrating the superiority of CMRCNet. CMRCNet not only achieves a higher similarity to the real spatial transcript expression levels but also exhibits significant potential in preserving biological heterogeneity and predicting tumor regions based on marker genes.

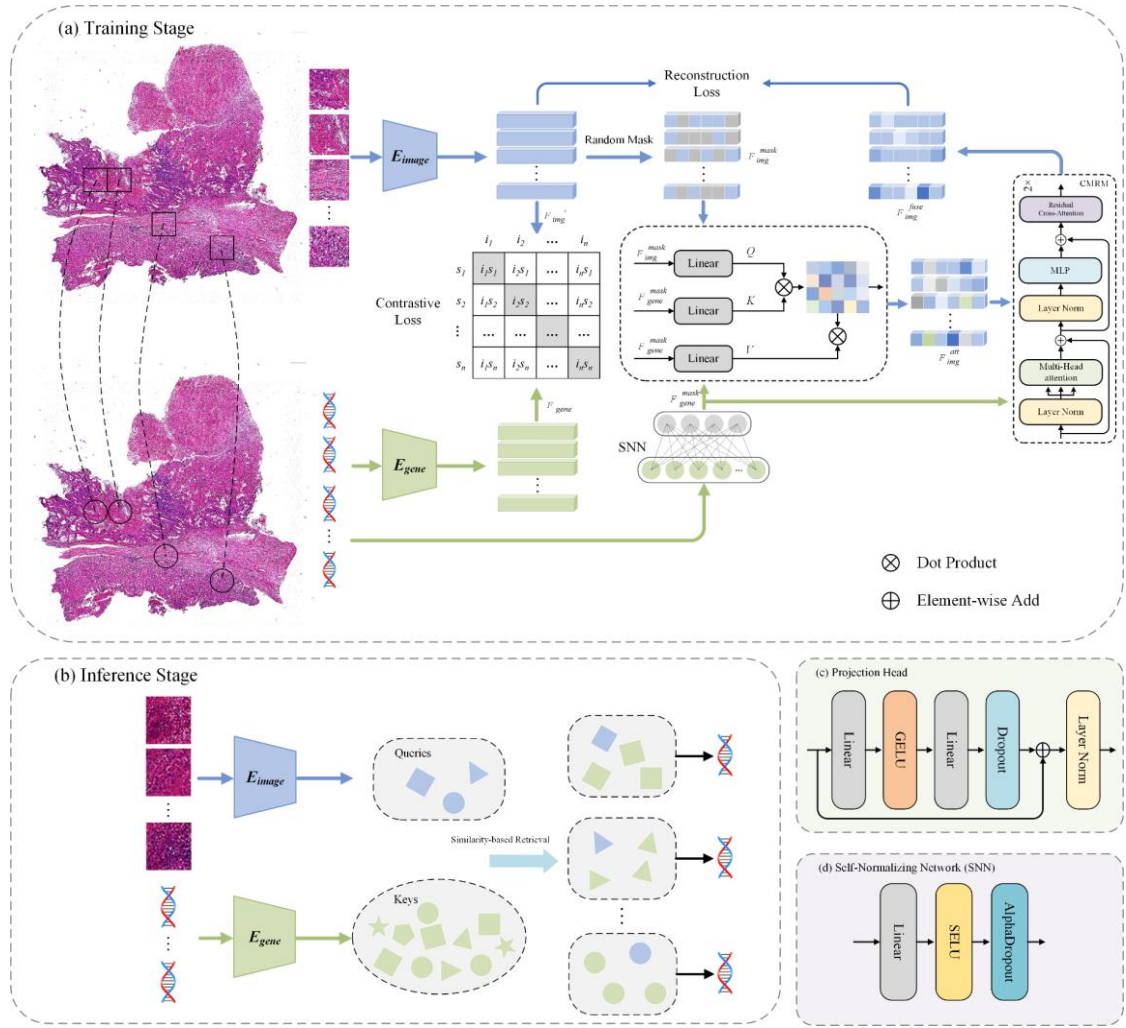

**Fig. 1 Overview of CMRCNet.** (a) Training Stage: The input data for CMRCNet in the training stage are histopathological patches and transcript gene expression, including contrastive loss and reconstruction loss. (b) Inference Stage: The input data for CMRCNet in the inference stage are histopathological patches from the test set and transcript gene expression from the train set, and the prediction results are obtained by similarity-based retrieval. (c) Projection Head is used to map high-dimensional features to lower dimensions for contrastive learning. (d) Self-normalizing network is used to map spatial transcriptomics data to lower dimensional features for constructing reconstruction loss.

**Results**

We compare CMRCNet with STNet[23], HisToGene[24], and the latest contrastive learning-based method, BLEEP[28]. Additionally, we also compare CMRCNet with cTranPath[31], a pioneering work of histopathology foundation model, whose pre-training datasets share the most similar cancer groups to the datasets we used. To avoid the impact of samples with different prediction difficulties on the method's evaluation, we employ k-fold cross-validation. The number of folds is equal to the number of samples in the dataset, such that that each sample participates at some point as a test sample in the evaluation of the method.

**Comparison of highly variable genes, highly expressed genes, and marker genes**

Table 1 presents the prediction results for the top 50 highly variable genes (HVGs), top 50 highly expressed genes (HEGs) among HVGs, and marker genes (MGs) used for analyzing disease progression across six datasets. All results are evaluated by PCC. The contrastive learning-based methods BLEEP and CMRCNet-50 are used, with 50 similar embeddings selected. For CMRCNet-500, 500 similar embeddings are selected. On all datasets, CMRCNet demonstrates optimal or second-best results.

For the prediction of HVGs and HEGs, CMRCNet-50 shows the most significant improvement on the COAD dataset. Compared to the end-to-end prediction model STNet, CMRCNet-50 improves HEGs prediction by 25.71% (PCC: 0.3872 vs 0.3080) and HVGs prediction by 24.12% (PCC: 0.4611 vs 0.3715). Compared to BLEEP, CMRCNet-50 shows an improvement of 10.34% (PCC: 0.3872 vs 0.3509) for HEGs and 10.18% (PCC: 0.4611 vs 0.4185) for HVGs. For the SKCM dataset, CMRCNet-50 achieves PCC values of 0.3420 for HEGs and 0.1966 for HVGs, representing the current best performance.

In the prediction of marker genes, CMRCNet-50 shows a significant improvement. Compared to BLEEP, CMRCNet-50 achieves a 40.76% improvement (PCC: 0.2704 vs 0.1921), and compared to STNet, it achieves a 330.57% improvement (PCC: 0.2704 vs 0.0628). The results of marker genes are shown in Supplementary Tables 1-6. Figure 2 presents box plots of the PCC scores for the top 50 HEGs and HVGs across the six datasets. The average and median PCC scores of CMRCNet-50 are significantly better than those of STNet and HisToGene, with positive PCC scores for most genes. Compared to BLEEP, CMRCNet-50 achieves higher median PCC scores across most samples, demonstrating stronger gene prediction capabilities.

To evaluate our method's ability to predict more genes, we present the prediction results for the top 150 HEGs, HVGs, and the top 250 HEGs, HVGs in Table 2. These genes may contain those that are significant for analyzing disease progression. For the prediction of the top 150 genes, we select 250 similar embeddings for the contrastive learning-based method. In the prediction of the top 250 genes, we select 500 similar embeddings. As shown in Table 2, our method achieves the best results on most datasets. Notably, all datasets show a similar trend, where predicting more genes typically results in a lower average PCC.

**Table 1. Results of comparison experiment (top 50 genes, Multi-fold Cross-Validation).** CMRCNet-50 selects 50 similar embeddings as keys for queries in the inference stage. CMRCNet-500 selects 500 similar embeddings as keys for queries in the inference stage. Best results are shown in red.

| Datasets | Models | HEG | HVG | Marker genes |
| --- | --- | --- | --- | --- |
| IDC-LymphNode | STNet | 0.2067 | 0.1806 | 0.0853 |
| | HisToGene | 0.2454 | 0.1922 | 0.0867 |
| | BLEEP | 0.2983 | 0.2179 | 0.1076 |
| | CMRCNet-50 | 0.3156 | 0.2345 | 0.1137 |
| | CMRCNet-500 | 0.3370 | 0.2479 | 0.1154 |
| SKCM | STNet | 0.1501 | 0.0825 | 0.0628 |
| | HisToGene | 0.2359 | 0.1319 | 0.0960 |
|         | BLEEP       | 0.3048  | 0.1809 | 0.1921 |
|         | CMRCNet-50  | 0.3420  | 0.1966 | 0.2704 |
|         | CMRCNet-500 | 0.3646  | 0.2268 | 0.2419 |
| READ    | STNet       | 0.2257  | 0.2538 | 0.1251 |
|         | HisToGene   | 0.1884  | 0.2136 | 0.1070 |
|         | BLEEP       | 0.3345  | 0.4222 | 0.2541 |
|         | CMRCNet-50  | 0.3327  | 0.4191 | 0.2505 |
|         | CMRCNet-500 | 0.3361  | 0.4317 | 0.2475 |
| COAD    | STNet       | 0.3080  | 0.3715 | 0.2221 |
|         | HisToGene   | 0.2905  | 0.3438 | 0.1046 |
|         | BLEEP       | 0.3509  | 0.4185 | 0.1454 |
|         | CMRCNet-50  | 0.3872  | 0.4611 | 0.1776 |
|         | CMRCNet-500 | 0.4478  | 0.5274 | 0.2136 |
| IDC     | STNet       | 0.2520  | 0.2204 | 0.2511 |
|         | HisToGene   | 0.2492  | 0.2210 | 0.2772 |
|         | BLEEP       | 0.2877  | 0.2424 | 0.2869 |
|         | CMRCNet-50  | 0.2993  | 0.2653 | 0.3108 |
|         | CMRCNet-500 | 0.3174  | 0.2838 | 0.3149 |
| PSC     | STNet       | 0.1817  | 0.2062 | 0.2382 |
|         | HisToGene   | 0.1757  | 0.1967 | 0.1489 |
|         | BLEEP       | 0.1837  | 0.2262 | 0.2934 |
|         | CMRCNet-50  | 0.1937  | 0.2352 | 0.3002 |
|         | CMRCNet-500 | 0.2010  | 0.2474 | 0.2990 |

**Table 2. Results of extended genes (Multi-fold Cross-Validation).** For the top 150 HEG and HVG predictions, CMRCNet selects 250 similar embeddings. For the top 250 HEG and HVG predictions, CMRCNet selects 500 similar embeddings. Best results are shown in red.

| Datasets      | Models    | HEG (top 150) | HVG (top 150) | HEG (top 250) | HVG (top 250) |
|---------------|-----------|---------------|---------------|---------------|---------------|
| IDC-LymphNode | STNet     | 0.1891        | 0.1819        | 0.1704        | 0.1690        |
|               | HisToGene | 0.2312        | 0.2131        | 0.2143        | 0.2042        |
|               | BLEEP     | 0.3036        | 0.2705        | 0.2892        | 0.2724        |
|               | CMRCNet   | 0.3138        | 0.2807        | 0.2983        | 0.2820        |
| SKCM          | STNet     | 0.1160        | 0.0971        | 0.0815        | 0.0852        |
|               | HisToGene | 0.1742        | 0.1460        | 0.1365        | 0.1315        |
|               | BLEEP     | 0.2517        | 0.2124        | 0.1989        | 0.1961        |
|               | CMRCNet   | 0.2642        | 0.2174        | 0.2086        | 0.2054        |
| READ          | STNet     | 0.1631        | 0.1910        | 0.1332        | 0.1548        |
|               | HisToGene | 0.1412        | 0.1637        | 0.1188        | 0.1345        |
|               | BLEEP     | 0.2694        | 0.3409        | 0.2326        | 0.2781        |
|               | CMRCNet   | 0.2641        | 0.3371        | 0.2268        | 0.2738        |
| COAD          | STNet     | 0.2855        | 0.3100        | 0.2610        | 0.2646        |
|               | HisToGene | 0.2578        | 0.2562        | 0.2189        | 0.2173        |
|               | BLEEP     | 0.3510        | 0.3617        | 0.3317        | 0.3350        |
|               | CMRCNet   | 0.3709        | 0.3841        | 0.3476        | 0.3494        |
| IDC | STNet | 0.2275 | 0.2037 | 0.1827 | 0.1766 |
| | HisToGene | 0.2462 | 0.2064 | 0.1916 | 0.1828 |
| | BLEEP | 0.3012 | 0.2540 | 0.2526 | 0.2428 |
| | CMRCNet | <span style="color:red">0.3156</span> | <span style="color:red">0.2759</span> | <span style="color:red">0.2678</span> | <span style="color:red">0.2589</span> |
| PSC | STNet | 0.1049 | 0.1083 | 0.0754 | 0.0804 |
| | HisToGene | 0.0977 | 0.0983 | 0.0695 | 0.0726 |
| | BLEEP | 0.1235 | 0.1296 | 0.0967 | 0.1034 |
| | CMRCNet | <span style="color:red">0.1288</span> | <span style="color:red">0.1343</span> | <span style="color:red">0.1007</span> | <span style="color:red">0.1075</span> |

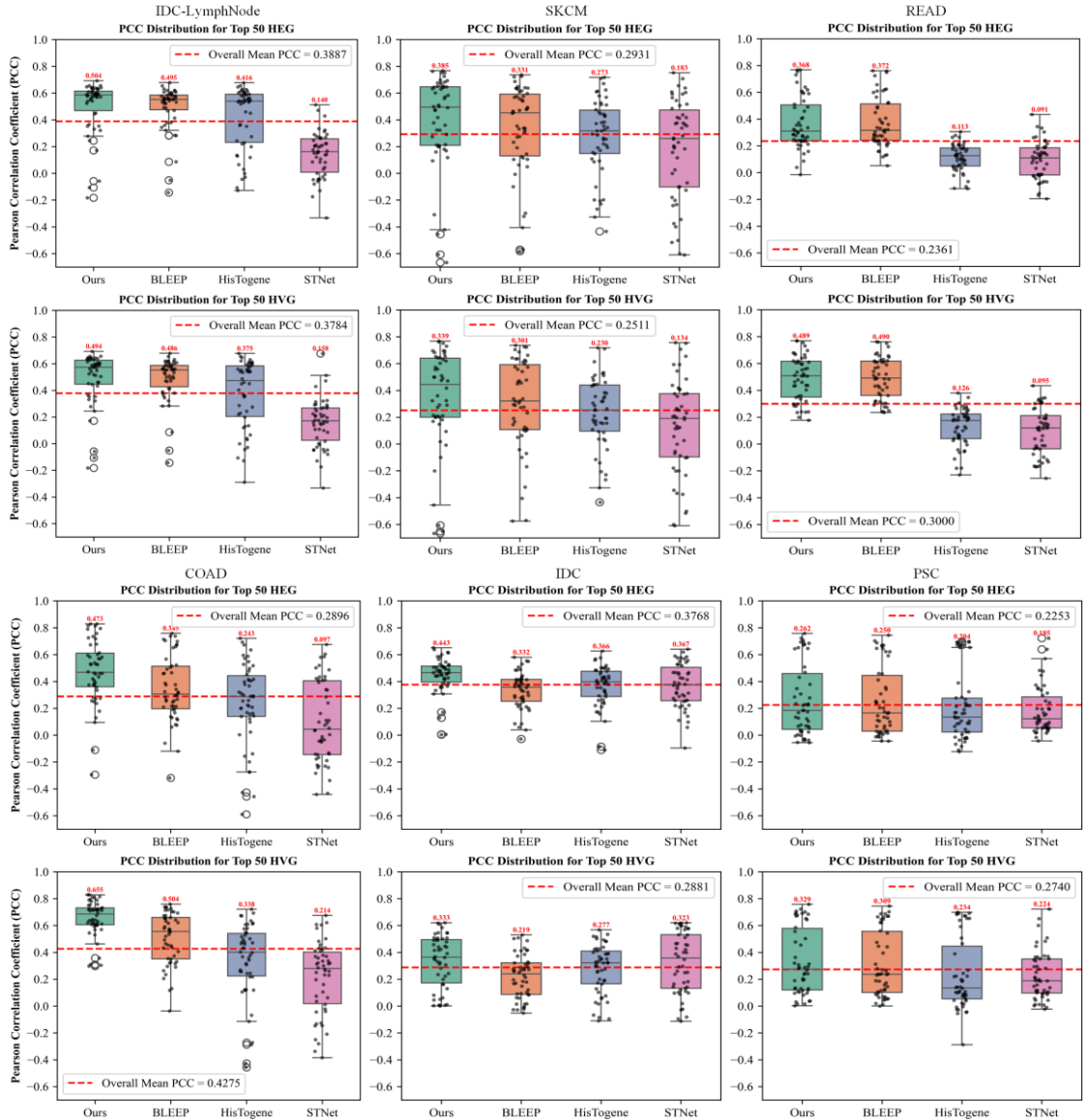

**Fig. 2 Boxplot of gene PCC distribution.** It reflects the predicted PCC distribution of the top 50 HEG and top 50 HVG for 6 samples from different diseases. Our method is evaluated with 50 similar embeddings selected. Green represents our method (CMRCNet), orange represents BLEEP, blue represents HisToGene, and pink represents STNet.

**Comparison of predicted and reference expression profiles**

To investigate the ability of different methods to predict gene expression levels across all genes in the samples,

Figure 3 compares the distribution of predicted expression profiles with reference expression profiles. The blue points represent the true values, while the yellow points represent the predicted values. The first row shows the average expression levels of different genes across the WSI, while the second row illustrates the variation in gene expression across different locations. In terms of predicting average expression levels, the points from STNet and HisToGene are the most scattered, indicating that it difficult to accurately predict gene expression levels. Compared to them, BLEEP and CMRCNet-50 demonstrate excellent performance on both low-expression and high-expression genes. For genes with median expression levels, CMRCNet-50 is more robust than BLEEP. In predicting gene expression variations, CMRCNet-50 shows slight improvements over other methods but still struggles to capture variations accurately.

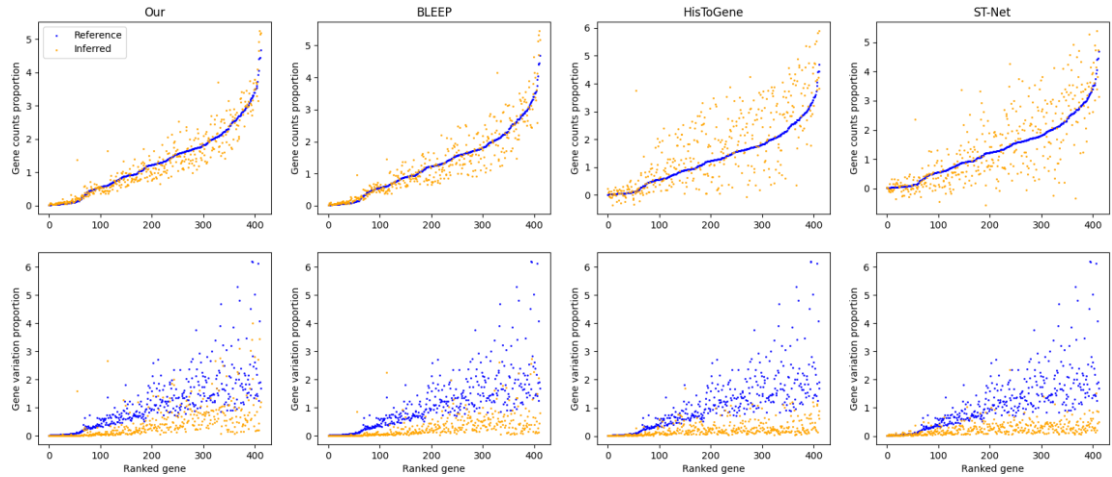

**Fig. 3 Comparison of predicted and reference expression profiles.** The blue points represent the distribution of ground truth and the yellow points represent the distribution of predictions. The first row represents the average expression of all genes in WSI. The second row represents the variation of all genes at different positions. Our method is evaluated with 50 similar embeddings selected.

**Analysis of gene-gene correlations**

Figure 4 shows the heatmap of gene-gene correlations, which reflects the ability of different methods to preserve biological heterogeneity. We select the top 50 HEGs and top 50 HVGs for testing. Hierarchical clustering is applied to classify and reorder these genes, with genes exhibiting similar expression patterns positioned closer together on the heatmap axes. In the heatmap of STNet's predicted results, the color blocks are scattered, indicating that the correlation between genes is not well preserved. Compared to STNet, HisToGene can recognize partial correlations between genes. Both BLEEP and our method demonstrate good performance in preserving biological heterogeneity. Regions corresponding to highly correlated gene subsets in the heatmap display similar patterns. Compared to BLEEP, our method better preserves the correlation in gene subsets that show negative correlations, represented by the blue areas in Figure 3.

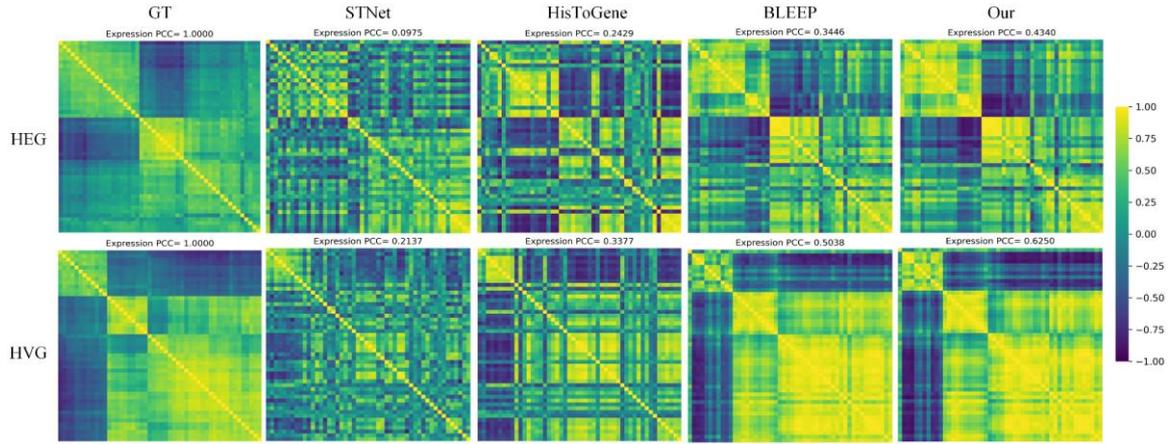

**Fig. 4 Heatmap of gene-gene correlations.** The first row represents the gene-gene correlations of the top 50 HEGs. The second row represents gene-gene correlations of the top 50 HVGs. Our method is evaluated with 50 similar embeddings selected.

**Analysis of highly related genes**

Table 3 presents the top 5 highly correlated genes from different methods across the 6 datasets. These genes show the high PCC scores in the predicted results. In the READ[32] and PSC[33] datasets, the most accurately predicted genes are relatively consistent across different methods. In the LYMPH_IDC[34], SKCM[26], and COAD[35] datasets, some genes show consistently high correlation prediction performance. Compared to STNet[23], HisToGene[24], and BLEEP[28], CMRCNet-50 shows a higher correlation with the true values for the prediction of these genes.

Figures 5, 6, and 7 show the distribution of expression levels for ITLN1, CD24, and GATA3 in colon adenocarcinoma, rectal adenocarcinoma tissue slides, and invasive ductal carcinoma tissue slides, respectively. Among them, ITLN1[36] is an independent favorable prognostic factor in colon adenocarcinoma, with multiple functions, including a role in inhibiting tumor angiogenesis. CD24[37] can be used to identify colorectal cancer stem cells, which are associated with tumor formation, chemotherapy resistance, and metastatic disease. GATA3 is a transcription factor that can be detected in cell nuclei of normal and neoplastic epithelial cells in the breast, that is used as lineage- and differentiation marker in IDC. High nuclear GATA3 expression has been shown to be associated with a favorable prognosis in certain subtypes of breast cancer. For ITLN1 prediction, BLEEP and CMRCNet-50 show higher similarity with the true values, with CMRCNet-50 achieving better PCC coefficients. For CD24 prediction, STNet and BLEEP only show high similarity in localized regions, while HisToGene performs better in low-expression regions but struggles with predicting high-expression regions of CD24. In comparison, CMRCNet-50 better captures the spatial variation of CD24 expression levels. For the prediction of GATA3, all four methods achieve comparable results. STNet and HisToGene achieve smoother results of the expression level. The contrast between the prediction results of BLEEP and our method in the region of different expression levels is more obvious. The original WSIs of these samples are shown in Supplementary Figures 1-3. Predicted PCCs of the top 50 HEGs and HVGs in these samples are shown in Supplementary Figures 4-6.

**Table 3. Results of top 5 correlated genes.** Red: Genes showing a high correlation in all four methods. Blue: Genes showing a high correlation in three methods.

|  | STNet |  | HisToGene |  | BLEEP |  | CMRCNet |  |
|---|---|---|---|---|---|---|---|---|
| Datasets | Gene | PCC | Gene | PCC | Gene | PCC | Gene | PCC |
| IDC-LymphNode | FABP4 | 0.4154 | HSP90AA1 | 0.4029 | UBA52 | 0.4851 | UBA52 | 0.5168 |
|  | PLIN1 | 0.3570 | FTL | 0.3828 | TMA7 | 0.4652 | MYL6 | 0.4804 |
| | HSP90AA1 | 0.3499 | MYL6 | 0.3809 | MYL6 | 0.4475 | TMA7 | 0.4778 |
| | KIAA1324 | 0.3480 | UBA52 | 0.3757 | HMGN2 | 0.4350 | SRSF3 | 0.4491 |
| | PLIN4 | 0.3449 | PPIB | 0.3635 | SRSF3 | 0.4337 | HMGN2 | 0.4454 |
| SKCM | MLANA | 0.5956 | MKI67 | 0.5972 | MKI67 | 0.6668 | MKI67 | 0.6845 |
| | S100A13 | 0.5880 | NSG1 | 0.5797 | SLC25A39 | 0.6414 | S100A13 | 0.6446 |
| | MKI67 | 0.5728 | STRADB | 0.5693 | NSG1 | 0.6413 | NSG1 | 0.6402 |
| | MITF | 0.5547 | SFRP1 | 0.5335 | TMEM150C | 0.6374 | SLC25A39 | 0.6393 |
| | C1QA | 0.5347 | TMEM150C | 0.5192 | S100A13 | 0.6314 | STRADB | 0.6260 |
| READ | ZG16 | 0.4074 | ITLN1 | 0.3499 | FCGBP | 0.5361 | FCGBP | 0.5375 |
| | ITLN1 | 0.3963 | COL1A2 | 0.3427 | MUC2 | 0.5105 | MUC2 | 0.5103 |
| | MGP | 0.3900 | MGP | 0.3122 | SPINK4 | 0.4976 | ITLN1 | 0.5008 |
| | SPINK4 | 0.3876 | SPINK4 | 0.3075 | ITLN1 | 0.4970 | SPINK4 | 0.4914 |
| | MALAT1 | 0.3753 | MYL9 | 0.3042 | ZG16 | 0.4913 | PIGR | 0.4879 |
| COAD | DPYSL3 | 0.7396 | CD24 | 0.6798 | CD24 | 0.6488 | CD24 | 0.6945 |
| | MYH14 | 0.6580 | DPYSL3 | 0.6243 | TAGLN | 0.6224 | TAGLN | 0.6607 |
| | CEACAM5 | 0.6347 | MYH14 | 0.6067 | DPYSL3 | 0.6106 | DPYSL3 | 0.6441 |
| | IGFBP7 | 0.6305 | SOX9 | 0.5958 | KRTCAP3 | 0.5874 | KRTCAP3 | 0.6306 |
| | RNF43 | 0.6247 | CEACAM5 | 0.5935 | SCNN1A | 0.5764 | PPP1R1B | 0.6173 |
| IDC | MMP2 | 0.4582 | SLC5A6 | 0.4647 | TFAP2A | 0.4799 | LYPD3 | 0.4721 |
| | TMEM147 | 0.4578 | OCIAD2 | 0.4625 | OCIAD2 | 0.4799 | FOXA1 | 0.4664 |
| | CD163 | 0.4328 | TMEM147 | 0.4381 | SEC24A | 0.4779 | TRAF4 | 0.4589 |
| | LYPD3 | 0.4291 | SRPK1 | 0.4291 | TRAF4 | 0.4684 | SEC11C | 0.4558 |
| | TACSTD2 | 0.4180 | BASP1 | 0.4067 | SLC5A6 | 0.4671 | SEC24A | 0.4533 |
| PSC | CYP3A4 | 0.6351 | CYP1A2 | 0.5351 | CYP3A4 | 0.6540 | CYP3A4 | 0.6551 |
| | CYP1A2 | 0.5482 | MT-CO3 | 0.5316 | CYP1A2 | 0.5881 | CYP1A2 | 0.5917 |
| | ORM1 | 0.5075 | CYP3A4 | 0.5225 | GLUL | 0.5505 | GLUL | 0.5470 |
| | GLUL | 0.4999 | MT-CO1 | 0.5222 | MT-CO3 | 0.5116 | MT-CO3 | 0.5227 |
| | MT-CO1 | 0.4642 | MT-CYB | 0.5199 | MT-CO1 | 0.4937 | MT-CO1 | 0.5105 |

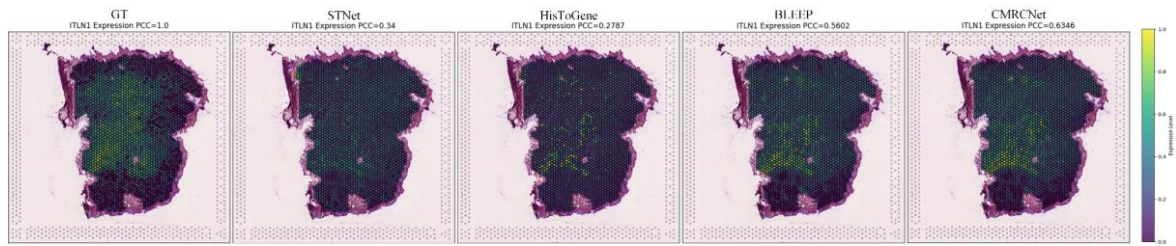

**Fig. 5 Distribution of the ITLN1 in rectal cancer tissue slices.**

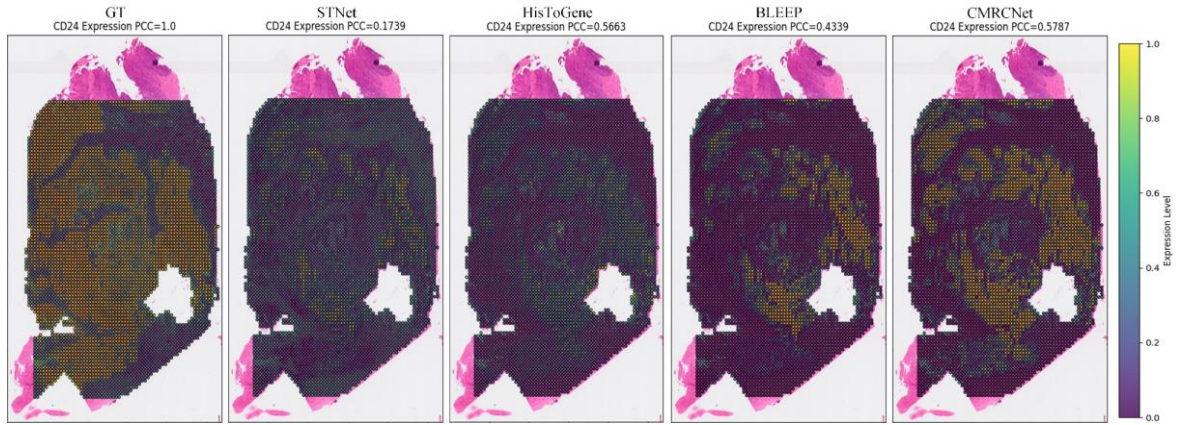

**Fig. 6 Distribution of the CD24 in colon cancer tissue slices.**

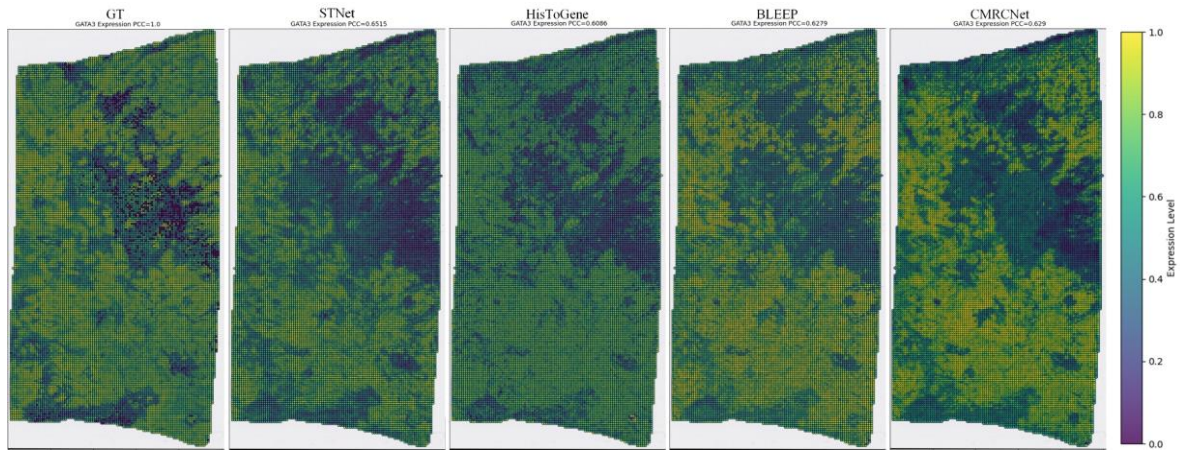

**Fig. 7 Distribution of the GATA3 in breast cancer tissue slices.**

**Comparison with pathology foundation model**

    Table 4 shows the comparison between CMRCNet and cTransPath. In the LYMPH_IDC, COAD, and IDC datasets, cTransPath shows better performance, consistent with the distribution of cancer types in the cTransPath training data. For the SKCM and READ datasets, CMRCNet achieves better prediction results. Even though cTransPath's pretrained data includes relevant disease samples, the inevitable class imbalance limits cTransPath's fine-tuning ability on the relevant datasets. For the PSC dataset, cTransPath does not show the ability to predict spatial transcriptomic expression. This is because there are no cancer samples in PSC, but liver inflammation data. Even though cTransPath's pretrained data includes over 500 liver WSIs, it still struggles to bridge the gap between different disease phenotypic patterns. In contrast, CMRCNet is suitable for different disease types and achieves better performance on the PSC dataset.

**Table 4. Results of comparison with pathology foundation model.** Best results are shown in red.

| Datasets | Models | HEG | HVG |
| --- | --- | --- | --- |
| IDC-LymphNode | cTranPath | 0.4326 | 0.3729 |
|  | CMRCNet-500 | 0.3370 | 0.2479 |
| SKCM | cTranPath | 0.3019 | 0.1841 |
|  | CMRCNet-500 | 0.3646 | 0.2268 |
| READ | cTranPath | 0.1892 | 0.3379 |
|  | CMRCNet-500 | 0.3361 | 0.4317 |

| | | | | |
|---|---|---|---|---|
| COAD | cTranPath | | 0.4982 | 0.6478 |
| | CMRCNet-500 | | 0.4478 | 0.5274 |
| IDC | cTranPath | | 0.4562 | 0.5139 |
| | CMRCNet-500 | | 0.3174 | 0.2838 |
| PSC | cTranPath | | 0.0169 | 0.0791 |
| | CMRCNet-500 | | 0.2010 | 0.2474 |

**Ablation experiment analysis**

To demonstrate the effectiveness of our method and the impact of its components, the ablation study results are presented in Table 5. We evaluate two types of reconstruction loss functions: cosine loss and MSE loss, three different masking rates: 0.3, 0.5, 0.7, and two types of reconstructed features. The number of similarity-based embeddings used during the inference stage is set to 50. Multi-fold cross-validation is conducted, and the results are reported as the average PCC scores across six datasets. When gene expression is predicted using contrastive learning alone, the PCC scores for HEG and HVG are 0.2933 and 0.2846, respectively. When image features are reconstructed using cosine loss as the reconstruction loss, the PCC scores under different masking rates are 0.3: (HEG: 0.3090, HVG: 0.3001), 0.5: (HEG: 0.3077, HVG: 0.2968), 0.7: (HEG: 0.3063, HVG: 0.2941). When the spatial transcriptomic expression is used as the reconstructed feature, the results are HEG: 0.3069, and HVG: 0.2946. When image features are reconstructed using MSE loss as the reconstruction loss, the PCC scores under different masking rates are 0.3: (HEG: 0.3020, HVG: 0.2921), 0.5: (HEG: 0.3117, HVG: 0.3020), 0.7: (HEG: 0.3006, HVG: 0.2924). When the spatial transcriptomic expression is used as the reconstructed feature, the results are HEG: 0.2996, and HVG: 0.2937. These results consistently outperform the PCC scores obtained using contrastive learning alone.

Table 5. Results of ablation experiments

| Loss Function | 0.3 | 0.5 | 0.7 | Reconstruction | HEG | HVG | Marker Genes |
|---|---|---|---|---|---|---|---|
| | | | | | 0.2933 | 0.2846 | 0.2132 |
| cosine | √ | | | Image | 0.3090 | 0.3001 | 0.2263 |
| cosine | | √ | | Image | 0.3077 | 0.2968 | 0.2205 |
| cosine | | | √ | Image | 0.3063 | 0.2941 | 0.2282 |
| cosine | | √ | | Spatial Transcription | 0.3069 | 0.2946 | 0.2116 |
| MSE | √ | | | Image | 0.3020 | 0.2921 | 0.2221 |
| MSE | | √ | | Image | 0.3117 | 0.3020 | 0.2372 |
| MSE | | | √ | Image | 0.3006 | 0.2924 | 0.2272 |
| MSE | | √ | | Spatial Transcription | 0.2996 | 0.2937 | 0.2190 |

**Discussion**

In this work, we propose a novel method (CMRCNet), which is based on contrastive learning, for predicting spatially resolved gene expression detected by spatial transcriptomics from routinely stained pathology images. During the training phase, contrastive learning is employed to minimize the distance between patches and their paired spatial transcriptomic expression in feature space. In the inference phase, CMRCNet selects gene expression embeddings that are similar in feature space as keys for each patch, and the average of the original expression levels of these keys is computed as the predicted result. Experiments on datasets from six different types of diseases demonstrate that CMRCNet outperforms others in spatial transcript expression prediction. It not only predicts gene spatial expression trends more accurately but also shows advantages in preserving biological heterogeneity.

CLIP[38] demonstrates remarkable potential in the image-text, providing a paradigm for cross-modal data alignment. BLEEP is the first to introduce CLIP to the task of spatial transcriptomic expression prediction by aligning histological images and spatial transcriptomics data through contrastive learning. However, previous studies show that contrastive learning, which relies solely on similarity-based determination, struggles to bridge the gap between multimodal data[29,39]. Numerous works propose various auxiliary tasks to further narrow the distance between images and text in semantic space. For instance, ALBEP[29] selects hard samples from a mini-batch and designs a classification task to optimize semantic matching. CE-CLIP[39] enhances CLIP's inference in visual-text combinations by directly generating hard samples. However, these methods are difficult to transfer to histopathology images and spatial transcriptomics data for two main reasons. First, text data contains rich and well-defined semantic information. Nouns in text represent the subjects in images, and adjectives reveal the characteristics of these subjects. For this reason, these methods[29,39] optimize the semantic matching process of image-text pairs. In contrast, spatial transcriptomic data lacks explicit semantic information, and the associations between different genes are complex. Moreover, noise is prevalent in the data, which affects the modeling of image-gene pairs. Second, a large number of image-text pairs are readily available for model training. CLIP is trained on 400 million image-text pairs. Thanks to the rich semantic information and vast training data, CLIP achieves success in the image and text domains. However, this success is difficult to replicate in the context of pathology patches and spatial transcriptomics. Collecting a large amount of spatial transcriptomic data for training is challenging, and further limitations arise from the different disease representation patterns and gene lengths.

    We propose an improved version of CLIP tailored for histopathology images and spatial transcriptomics data, designed for gene expression prediction from images. Unlike methods that rely on semantic information, our method operates based on feature pixels. It incorporates two loss functions: a contrastive learning loss and a reconstruction loss. The contrastive learning loss establishes associations between paired cross-modal data in the feature space through similarity computation, while the reconstruction loss directly integrates multimodal data at the feature level, facilitating cross-modal interaction learning in the encoder. Furthermore, our approach enables training and inference on small-scale datasets. Compared to end-to-end prediction models, it offers greater flexibility during inference, as it does not require fixed labels. Most end-to-end methods focus on predicting a fixed number of highly variable genes within a dataset[23,24,40]. However, due to expression heterogeneity across samples and the fact that certain marker genes do not exhibit strong variability, these important genes are often overlooked. Our method accounts for sample-specific differences during inference, resulting in greater generalizability. To further analyze the predictive capability of CMRCNet, we compare CMRCNet with the pathology foundation model cTransPath. cTransPath is trained on over 35,000 WSIs, totaling approximately 15 million patches, and includes multiple cancer types. The most common tissue types in these datasets are breast tissue, colon tissue, kidney tissue, and lung tissue, as common cancers often occur in these tissues. Compared to cTransPath, our image encoder uses the VIT-B version, which is trained only on the datasets we used, without incorporating any pretrained knowledge. In addition, we conduct more comprehensive and reliable validation. By evaluating six different disease datasets and employing multi-fold cross-validation, we mitigate the randomness introduced by varying levels of prediction difficulty across samples within the datasets.

    In the ablation study, the impact of the cross-modal reconstruction loss is highly significant. Overall, across different experimental settings, the results of CMRCNet consistently outperform models that rely solely on contrastive learning loss. Specifically, when using the cosine similarity loss for model optimization, we recommend selecting a lower mask rate. The cosine similarity loss reflects the similarity between feature vectors in the embedding space, and preserving more original information is beneficial for model learning. When optimizing the model with MSE loss, an appropriate mask rate is more critical, which in our task is 0.5. MSE focuses more on feature pixels, where a smaller masking ratio may fail to fully activate the model's potential, while a higher ratio may introduce learning difficulties. Additionally, under the same loss function and mask rate, reconstructing image

features proves to be the better choice. This is due to, on the one hand, spatial transcriptomic data inherently containing noise, which unavoidably interferes with the reconstruction process. On the other hand, the image encoder exhibits a greater capacity for feature extraction and representation learning.

A major limitation of our method lies in inference speed, which stems from its unique design paradigm. End-to-end models directly infer gene expression levels from histopathology images. In contrast, our method requires selecting similar transcriptomic embeddings for each patch in the test cohort. Consequently, all samples in the dataset participate in the computation of the similarity matrix. As a result, in addition to being influenced by the complexity of the encoder, the inference speed is also dependent on the number of samples in the dataset. Future work can be explored in two directions. (1) Our method has the potential to be transferred to other tasks. In this study, CMRCNet demonstrates promising performance in predicting marker genes, as shown in Figures 5–7. The expression of these genes can be used to distinguish relevant parts of cancerous tissues from normal tissues (e.g., presence of CD27 or absence of ITLN1), or to identify cells with a particular type of differentiation or histogenetic "lineage" (e.g., GATA3). Furthermore, we establish correlations between histopathology images and transcriptomic data through contrastive learning. A multimodal knowledge-embedded encoder can leverage knowledge distillation techniques to transfer these correlations to tasks such as survival prediction and cancer subtype classification, thereby enhancing performance under a single modality. (2) Our method can be extended to a wide range of different markers in other data sets and may also be used in other types of spatially resolved image data. Future applications of the method in real-world clinical settings may include computer-based prediction of biomarker expression on routinely stained tissues to guide subsequent biomarker analysis, for example to select appropriate marker panels, or to save time and accelerate case management (e.g., initiate clinical staging) even before the more time-consuming standard biomarker analysis is completed. Since the introduction of CLIP, numerous CLIP-based methods have been developed to model the correlations between medical images and clinical text reports. However, very few universal methods have been proposed for structured clinical data, such as spatial transcriptomic expression, radiomic features, and serum antigen expression levels. In the future, we will validate our model on further types of medical data.

## Method

**Datasets**

To validate the effectiveness of the proposed method, we train and evaluate CMRCNet on six datasets reflecting different diseases, including axillary lymph nodes in invasive ductal carcinoma (IDC-LymphNode)[34], skin cutaneous melanoma (SKCM)[26], rectal adenocarcinoma (READ)[32], colonic adenocarcinoma (COAD)[35], invasive ductal carcinoma (IDC)[41], and primary sclerosing cholangitis (PSC)[33]. All these datasets are collected and curated by the HEST-benchmark[26]: https://huggingface.co/datasets/MahmoodLab/hest.

IDC-LymphNode[34]: This dataset consists of four whole-slide images (WSIs) and spatial transcriptomics samples from patients diagnosed with breast cancer. The spatial transcriptomics data are obtained using the Visium transcriptomics technology[42], containing a total of 19,964 transcriptomic spots and expression levels of 33,921 genes. Scanpy[43] is used to select the top 1,000 highly variable genes from each sample, resulting in a final dataset comprising 3,467 genes. For the evaluation of marker genes and genes influencing disease progression, the following genes are included in our study: ZNF276, ZNF750, YBX1, VHL, RUNX2, RND, ERCC4, PADI2, S100A8, and S100A6.

SKCM[26]: This dataset includes samples from two patients with skin cancer. The spatial transcriptomic data are obtained using Xenium technology, which contains 5,716 transcriptomic spots and the expression levels of 541 genes. All of these genes are included in our study. We select MIA, PMEL, MLANA, MITF, and TYR as the marker genes for the SKCM dataset.

READ[32]: This dataset consists of four samples from two patients with rectal adenocarcinoma. The spatial transcriptomic data are obtained using Visium, containing 8,407 transcriptomic spots and expression levels of 36,601 genes. Similar to the preprocessing steps of the IDC-LymphNodes dataset, we select 3,355 highly variable genes. The marker genes for the READ dataset include: OLFM4, PLA2G2A, PLA2G2D, MMP1, MMP11, MMP12, SMAD4, MUC13, SLC26A3, ZG16 and NDRG1.

COAD[35]: This dataset includes four WSIs and spatial transcript samples from patients with colonic adenocarcinoma. The spatial transcriptomic data are obtained using Xenium technology, comprising 18,523 transcriptomic spots and the expression levels of 541 genes. After excluding genes with different names and blank-named genes, 412 genes are included in our study. The marker genes for the COAD dataset include CLCA1, CD8A, PRF1, GZMA, and GZMK.

IDC[41]: This dataset includes four samples from patients with invasive ductal carcinoma. The spatial transcriptomic data are obtained using Xenium technology, comprising 44,974 transcriptomic spots and the expression levels of 541 genes. After filtering out genes with different names and blank-named genes, 500 genes are included in our study. The marker genes for the IDC dataset include MKI67, ERBB2, PRF1, ESR1, PGR, GATA3, AR and EGFR.

PSC[33]: This dataset includes four samples from patients with Primary Sclerosing Cholangitis. The spatial transcriptomic data are obtained using Visium technology, comprising 19,968 transcriptomic spots and the expression levels of 36,601 genes. Using Scanpy, the top 1,000 highly variable genes are selected for each sample, and 3,355 genes are included in the study. The marker genes for PSC include CYP3A4, VWF, KRT7, ACTA2, and DCN.

**Evaluation metrics**

We use the Pearson Correlation Coefficient (PCC) to evaluate our method, which is the most commonly evaluated metric for gene expression prediction[24,28,40]. The PCC reflects the linear relationship between the predicted values and the true gene expression levels, with a range of [-1, 1]. In spatial transcriptomic gene prediction, PCC reflects the consistency between the prediction and ground truth in spatial variation patterns. The PCC is calculated as follows:

$$r = \frac{\sum_{i=1}^{n}(X_i - \bar{X})(Y_i - \bar{Y})}{\sqrt{\sum_{i=1}^{n}(X_i - \bar{X})^2}\sqrt{\sum_{i=1}^{n}(Y_i - \bar{Y})^2}},$$

where $X$ and $Y$ represent the two variables involved in the computation, $i$ denotes the index of the data points, and $n$ represents the total number of data points. $\bar{X}$ and $\bar{Y}$ denote the mean values of the two variables, respectively.

**Implementation**

Our framework is implemented using PyTorch. Model training and evaluation are performed on a workstation equipped with an Nvidia RTX A5000 24G GPU. All WSIs are divided into patches with a resolution of 224×224. We preprocess the gene expression data using total counts normalization and log transformation to eliminate the effects of sequencing depth. During the training phase, data augmentation techniques include random horizontal flipping, random vertical flipping, and random rotation. We use the AdamW optimizer[44] with a learning rate of 0.0001 to train our model for 60 epochs, with a batch size of 64.

**CMRCNet architecture**

Our novel method CMRCNet is inspired by CLIP[38] and BLEEP[28]. However, CLIP is primarily designed for large-scale contrastive learning between images and text, while BLEEP focuses solely on narrowing the spatial gap between pathological morphology and spatial gene expression, without enabling deeper cross-modal interaction. In contrast, CMRCNet utilizes cross-modal reconstruction to achieve interaction and alignment of histopathological morphology and gene expression levels at the feature pixel level, and is suitable for small-scale datasets. The overall workflow of CMRC-Net is illustrated in Figure 1. It consists of three main stages: contrastive learning, cross-modal reconstruction, and similarity-based retrieval during inference. During the training phase, the model takes paired histopathology image patches and spatial transcriptomic expression profiles as input. In the inference phase, histopathology images from the test set are fed into the image encoder, while spatial transcriptomic data from the training set serve as input to the gene encoder. The final spatial transcriptomic expression predictions are obtained through similarity-based retrieval.

**Contrastive learning**

Each whole slide image $X \in \mathbb{R}^{H \times W \times 3}$ is first divided into patches $I \in \mathbb{R}^{N \times L \times L \times 3}$, where $N$ represents the number of patches, and $L$ denotes the patch resolution. In contrastive learning, the inputs for the image encoder and transcriptomic encoder are $I \in \mathbb{R}^{B \times L \times L \times 3}$ and $S \in \mathbb{R}^{B \times d}$, where $B$ represents the batch size, set to 64 in our method, and $d$ denotes the dimension of gene expression in spatial transcriptomics. To ensure consistency between the architectures of the image encoder in contrastive learning and the cross-attention module in cross-modal reconstruction, we employ the Vision Transformer (ViT) base version as the image encoder. After feature extraction, the output of the image encoder is $F_{img} \in \mathbb{R}^{B \times 50 \times 768}$, where 50 represents the number of tokens, and 768 is the embedding dimension of ViT-B. Next, the average value of $F_{img}$ is computed along the token dimension and fed into the projection head, which maps the high-dimensional hidden representations from the encoder to a lower-dimensional space. The projection head consists of two fully connected layers, a GELU activation function, dropout, and layer normalization, which is consistent with the design in BLEEP. The architecture of the projection head is shown in Figure 1(c). The output of the projection head is $F'_{img} \in \mathbb{R}^{B \times 256}$, which serves as the feature representation for contrastive learning. For spatial transcriptomic expression $S \in \mathbb{R}^{B \times d}$, the projection head directly acts as a gene encoder, extracting features from the expression data, with the output given by $F_{gene} \in \mathbb{R}^{B \times 256}$.

We follow the construction of the contrastive loss as implemented in BLEEP[28], where similarity computation and label construction are performed within a mini-batch. First, the intra-modal similarity is computed, as shown in the following equation:

$$sim(F'_{img}, F'_{img}) \in \mathbb{R}^{B \times B} = F'_{img} \cdot {F'_{img}}^T$$

$$sim(F_{gene}, F_{gene}) \in \mathbb{R}^{B \times B} = F_{gene} \cdot {F_{gene}}^T$$

Next, the cross-modal similarity is computed, as shown in the following equation:

$$sim(F'_{img}, F_{gene}) \in \mathbb{R}^{B \times B} = F'_{img} \cdot {F_{gene}}^T$$

$$sim(F_{gene}, F'_{img}) \in \mathbb{R}^{B \times B} = F_{gene} \cdot {F'_{img}}^T$$

The label construction process for contrastive learning is shown in the following equation:

$$target = softmax(\frac{sim(F'_{img}, F'_{img}) + sim(F_{gene}, F_{gene})}{2} \times \tau)$$

Where $\tau$ is the temperature function in contrastive learning. We follow the settings from BLEEP, where $\tau = 1.0$. Cross-entropy (CE) is used as the contrastive learning loss function, modeling the correlation between the two modalities based on similarity calculation. The overall contrastive learning loss function is defined as follows:

$$Loss_c = CE(sim(F'_{img}, F_{gene}), target) + CE(sim(F_{gene}, F'_{img}), target^T)$$

**Cross-modal reconstruction**

Thanks to the rich semantic information and large training datasets, CLIP[38] and its improved versions have become a paradigm for image-text contrastive learning. However, these architectures are difficult to transfer to image-spatial transcriptomics due to the lack of clear semantic information in spatial transcriptomic data and the unavoidable limitations in the data scale. To enable the encoder to further learn cross-modal interactions and promote the fusion of the two modalities, inspired by MAE[45], we propose a cross-modal reconstruction task as an auxiliary task to achieve feature-level fusion between tissue morphological features and gene expression levels. Unlike MAE which trains a robust image encoder by randomly masking the original image patches and utilizing unimodal self-reconstruction, our cross-modal reconstruction strategy works within multiple modalities and hidden layer feature embeddings, which is used to enhance the interaction and alignment of multimodal features.

For the output of the image encoder $F_{img} \in \mathbb{R}^{B \times 50 \times 768}$, we use a fully connected layer to map the embedding dimension from 768 to 256. For spatial transcriptomic data, we utilize Self-Normalizing Networks (SNN)[46] to map the transcript expression from the initial dimension to 2,048 dimensions, and then perform a dimensionality transformation to construct the transcriptomic features $F_{gene}^{mask} \in \mathbb{R}^{B \times 8 \times 256}$. The SNN, which is designed to encode gene expression sequences[47,48], consists of fully connected layers, a SELU activation function, and Alpha Dropout with a dropout probability of 0.25. The architecture of SNN is shown in Figure 1(d). The SELU is continuous and differentiable at all points, avoiding the issue of neuron death, and causes the output to tend towards zero mean and unit variance. Combined with SELU, Alpha Dropout further maintains the self-normalizing property of the output. First, we randomly mask 50% of the feature pixels in the image features, and the masked feature representation is denoted as $F_{img}^{mask} \in \mathbb{R}^{B \times 50 \times 256}$. In our ablation evaluation, the combination of a 50% mask rate and MSE reconstruction loss function shows the best performance. Next, the masked image features and transcriptomic features are input into the cross-attention module for preliminary reconstruction. Since image features serve as the main feature for reconstruction, in the preliminary reconstruction process, the Query ($Q$) is generated from $F_{img}^{mask}$, while the Key ($K$) and Value ($V$) are generated from $F_{gene}^{mask}$. The construction process of $Q$, $K$, and $V$ is as follows:

$$Q = F_{img}^{mask} \cdot W_Q$$

$$K = F_{gene}^{mask} \cdot W_K$$

$$V = F_{gene}^{mask} \cdot W_V$$

The output of the cross-attention is denoted as $F_{img}^{att} \in \mathbb{R}^{B \times 50 \times 256}$. This output is calculated as

$$F_{img}^{att} = softmax\left(\frac{QK^T}{\sqrt{d_k}}\right)V,$$

where $d_k$ represents the dimension of the feature embeddings. Cross-attention is implemented using the multi-head approach, with the number of heads set to 4. Next, $F_{img}^{att}$ is fed into two concatenated cross-modal fusion modules for feature extraction and further cross-modal reconstruction. Our cross-modal fusion module consists of two steps: image feature extraction and the fusion of image and gene expression features. To maintain consistency with the image encoder feature extraction, the cross-modal fusion module adopts the structure of Transformer blocks, which includes layer normalization, multi-head self-attention mechanism, and linear mapping layers. In the cross-modal fusion module, the number of heads is set to 8. A cross-attention module is added as the final layer of the cross-modal fusion module. The architecture of the cross-attention module is shown in Figure 8. Unlike cross-attention in the initial reconstruction, we also generate Value ($V$) from the image features and add it to the features weighted by the attention matrix in a residual-connected manner. This design has two motivations: 1. It prevents the information from overly favoring the spatial transcriptomic modality. Spatial transcriptomic data inevitably contains noise. When features overly rely on gene modality representations in the reconstruction process, noise is often retained in the fused features, making optimization difficult. 2. It preserves feature representations from the initial fusion and self-reconstruction. Before entering the cross-attention mechanism, image features undergo initial reconstruction and complete their information interaction within the Transformer block. The mappings from these features are added to the final feature using a residual connection. This ensures that the reconstruction is not solely dependent on the gene expression features. Unmasked tissue morphology features and fused features from each stage also contribute to the reconstruction process. The output of the cross-modal fusion module is denoted as $F_{fuse} \in \mathbb{R}^{B \times 50 \times 256}$.

In the cross-modal reconstruction task, we use the MSE loss function to guide the model's learning. The image feature embeddings used for contrastive learning serve as the reconstruction target. $F_{fuse}$ is averaged along the token dimension to produce the reconstruction result, denoted as $F_r \in \mathbb{R}^{B \times 256}$. The overall loss function for the cross-modal reconstruction is as follows:

$$Loss_r = MSE(F_r, F'_{img})$$

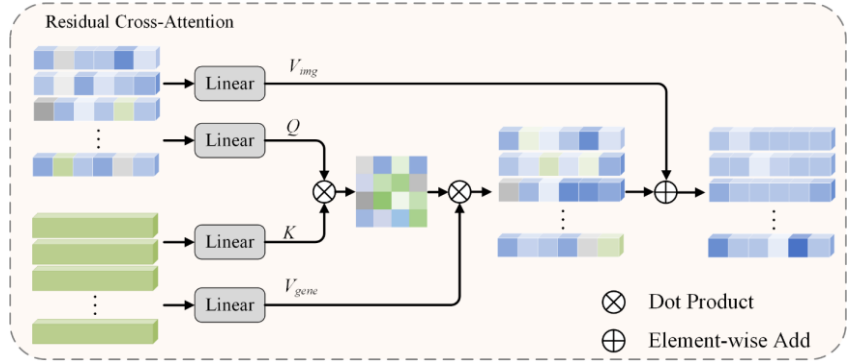

**Fig. 8 Overview of residual cross-attention module.** It is used to fuse histopathological features and gene expression features in the cross-modal reconstruction stage.

**Loss function**

Our method includes two loss functions: the contrastive learning loss and the cross-modal reconstruction loss. Since the two loss functions differ in magnitude, we use a logarithmic function to balance the optimization process and prevent one loss from dominating the model training. The overall loss function is as follows:

$$Loss_{total} = \log(1 + Loss_c) + \log(1 + Loss_r)$$

**Similarity-based retrieval in inference stage**

The inference of spatial transcriptomic expression from images is completed based on similarity retrieval. The process of inference is shown in Figure 1(b). During the inference stage, we only retain the VIT and projection head used for image encoding in contrastive learning, as well as the projection head used for spatial transcriptomic encoding. First, the WSIs from the test dataset are divided into patches based on the spatial transcriptomic location information and input into the VIT and Projection Head, producing image feature embeddings $F_{img}^{test} \in \mathbb{R}^{B \times 256}$. Next, the spatial transcriptomic data from the training dataset is used as input for the trained gene encoder, producing gene feature embeddings $F_{gene}^{test} \in \mathbb{R}^{M \times 256}$, where $M$ represents the total number of transcriptomic spots in the training dataset. Then, the image feature embeddings and gene feature embeddings are used to compute the cosine similarity, resulting in a similarity matrix $S \in \mathbb{R}^{B \times M}$. The top 50 gene feature embeddings most similar to the current image feature embedding are selected as keys. The average gene expression values of the corresponding transcriptomic spots are taken as the final prediction results.

**Reference**

1　　Bray, F. *et al.* Global cancer statistics 2022: GLOBOCAN estimates of incidence and mortality worldwide for 36 cancers in 185 countries. *CA: a cancer journal for clinicians* **74**, 229-263 (2024).

2　　Sung, H. *et al.* Global cancer statistics 2020: GLOBOCAN estimates of incidence and mortality worldwide for 36 cancers in 185 countries. *CA: a cancer journal for clinicians* **71**, 209-249 (2021).

3       Kleihues, P. *et al.* The WHO classification of tumors of the nervous system. *Journal of Neuropathology & Experimental Neurology* **61**, 215-225 (2002).

4       Brierley, J. D., Gospodarowicz, M. K. & Wittekind, C. *TNM classification of malignant tumours*.   (John Wiley & Sons, 2017).

5       Fu, Y. *et al.* Pan-cancer computational histopathology reveals mutations, tumor composition and prognosis. *Nature cancer* **1**, 800-810 (2020).

6       Swanton, C. *et al.* Embracing cancer complexity: Hallmarks of systemic disease. *Cell* **187**, 1589-1616 (2024).

7       Sparks, R. & Madabhushi, A. Explicit shape descriptors: Novel morphologic features for histopathology classification. *Medical image analysis* **17**, 997-1009 (2013).

8       Hölscher, D. L. *et al.* Next-Generation Morphometry for pathomics-data mining in histopathology. *Nature communications* **14**, 470 (2023).

9       Ektefaie, Y. *et al.* Integrative multiomics-histopathology analysis for breast cancer classification. *NPJ Breast Cancer* **7**, 147 (2021).

10      Hu, J. *et al.* SpaGCN: Integrating gene expression, spatial location and histology to identify spatial domains and spatially variable genes by graph convolutional network. *Nature methods* **18**, 1342-1351 (2021).

11      Rakha, E., Toss, M. & Quinn, C. Specific cell differentiation in breast cancer: a basis for histological classification. *Journal of clinical pathology* **75**, 76-84 (2022).

12      Takahara, T., Murase, Y. & Tsuzuki, T. Urothelial carcinoma: variant histology, molecular subtyping, and immunophenotyping significant for treatment outcomes. *Pathology* **53**, 56-66 (2021).

13      Tsuneki, M., Abe, M., Ichihara, S. & Kanavati, F. Inference of core needle biopsy whole slide images requiring definitive therapy for prostate cancer. *BMC cancer* **23**, 11 (2023).

14      Kilim, O. *et al.* Histopathology and proteomics are synergistic for high-grade serous ovarian cancer platinum response prediction. *npj Precision Oncology* **9**, 27 (2025).

15      Handley, K. F. *et al.* Classification of high-grade serous ovarian cancer using tumor morphologic characteristics. *JAMA network open* **5**, e2236626-e2236626 (2022).

16      Cannella, R. *et al.* Association of LI-RADS and histopathologic features with survival in patients with solitary resected hepatocellular carcinoma. *Radiology* **310**, e231160 (2024).

17      Ho, K. K. *et al.* A proposed clinical classification for pituitary neoplasms to guide therapy and prognosis. *The Lancet Diabetes & Endocrinology* **12**, 209-214 (2024).

18      Garcia-Alonso, L. *et al.* Mapping the temporal and spatial dynamics of the human endometrium in vivo and in vitro. *Nature genetics* **53**, 1698-1711 (2021).

19      Chen, W.-T. *et al.* Spatial transcriptomics and in situ sequencing to study Alzheimer's disease. *Cell* **182**, 976-991. e919 (2020).

20      Williams, C. G., Lee, H. J., Asatsuma, T., Vento-Tormo, R. & Haque, A. An introduction to spatial transcriptomics for biomedical research. *Genome Medicine* **14**, 68 (2022).

21      Jin, Y. *et al.* Advances in spatial transcriptomics and its applications in cancer research. *Molecular Cancer* **23**, 129 (2024).

22      Yu, Q., Jiang, M. & Wu, L. Spatial transcriptomics technology in cancer research. *Frontiers in Oncology* **12**, 1019111 (2022).

23      He, B. *et al.* Integrating spatial gene expression and breast tumour morphology via deep learning. *Nature biomedical engineering* **4**, 827-834 (2020).

24      Pang, M., Su, K. & Li, M. Leveraging information in spatial transcriptomics to predict super-resolution gene expression from histology images in tumors. *BioRxiv*, 2021.2011. 2028.470212 (2021).

25   Yang, Y., Hossain, M. Z., Stone, E. & Rahman, S. Spatial transcriptomics analysis of gene expression prediction using exemplar guided graph neural network. *Pattern Recognition* **145**, 109966 (2024).

26   Jaume, G. *et al.* Hest-1k: A dataset for spatial transcriptomics and histology image analysis. *Advances in Neural Information Processing Systems* **37**, 53798-53833 (2025).

27   Rahaman, M. M., Millar, E. K. & Meijering, E. Breast cancer histopathology image-based gene expression prediction using spatial transcriptomics data and deep learning. *Scientific Reports* **13**, 13604 (2023).

28   Xie, R. *et al.* Spatially Resolved Gene Expression Prediction from Histology Images via Bi-modal Contrastive Learning. *Advances in Neural Information Processing Systems* **36** (2024).

29   Li, J. *et al.* Align before fuse: Vision and language representation learning with momentum distillation. *Advances in neural information processing systems* **34**, 9694-9705 (2021).

30   Wu, Z., Xiong, Y., Yu, S. X. & Lin, D. Unsupervised feature learning via non-parametric instance discrimination in *Proceedings of the IEEE conference on computer vision and pattern recognition.* 3733-3742 (2018).

31   Wang, X. *et al.* Transformer-based unsupervised contrastive learning for histopathological image classification. *Medical image analysis* **81**, 102559 (2022).

32   Valdeolivas, A. *et al.* Charting the heterogeneity of colorectal cancer consensus molecular subtypes using spatial transcriptomics. *bioRxiv*, 2023.2001. 2023.525135 (2023).

33   Andrews, T. S. *et al.* Single-cell, single-nucleus, and spatial transcriptomics characterization of the immunological landscape in the healthy and PSC human liver. *Journal of Hepatology* **80**, 730-743 (2024).

34   Liu, T. *et al.* Single cell profiling of primary and paired metastatic lymph node tumors in breast cancer patients. *Nature Communications* **13**, 6823 (2022).

35   Oliveira, M. F. *et al.* Characterization of immune cell populations in the tumor microenvironment of colorectal cancer using high definition spatial profiling. *BioRxiv*, 2024.2006. 2004.597233 (2024).

36   Chen, L. *et al.* ITLN1 inhibits tumor neovascularization and myeloid derived suppressor cells accumulation in colorectal carcinoma. *Oncogene* **40**, 5925-5937 (2021).

37   Huang, J. L., Oshi, M., Endo, I. & Takabe, K. Clinical relevance of stem cell surface markers CD133, CD24, and CD44 in colorectal cancer. *American Journal of Cancer Research* **11**, 5141 (2021).

38   Radford, A. *et al.* Learning transferable visual models from natural language supervision in *International conference on machine learning.* 8748-8763 (2021).

39   Zhang, L., Awal, R. & Agrawal, A. Contrasting intra-modal and ranking cross-modal hard negatives to enhance visio-linguistic compositional understanding in *Proceedings of the IEEE/CVF Conference on Computer Vision and Pattern Recognition.* 13774-13784 (2024).

40   Jia, Y., Liu, J., Chen, L., Zhao, T. & Wang, Y. THItoGene: a deep learning method for predicting spatial transcriptomics from histological images. *Briefings in Bioinformatics* **25**, bbad464 (2024).

41   Janesick, A. *et al.* High resolution mapping of the tumor microenvironment using integrated single-cell, spatial and in situ analysis. *Nature communications* **14**, 8353 (2023).

42   Ståhl, P. L. *et al.* Visualization and analysis of gene expression in tissue sections by spatial transcriptomics. *Science* **353**, 78-82 (2016).

43   Wolf, F. A., Angerer, P. & Theis, F. J. SCANPY: large-scale single-cell gene expression data analysis. *Genome biology* **19**, 1-5 (2018).

44   Loshchilov, I. & Hutter, F. Decoupled weight decay regularization. *arXiv preprint arXiv:1711.05101* (2017).

45   He, K. *et al.* Masked autoencoders are scalable vision learners in *Proceedings of the IEEE/CVF conference*

*on computer vision and pattern recognition.* 16000-16009 (2022).

46    Klambauer, G., Unterthiner, T., Mayr, A. & Hochreiter, S. Self-normalizing neural networks. *Advances in neural information processing systems* **30** (2017).

47    Ding, K., Zhou, M., Metaxas, D. N. & Zhang, S. Pathology-and-genomics multimodal transformer for survival outcome prediction in *International Conference on Medical Image Computing and Computer-Assisted Intervention.* 622-631 (2023).

48    Chen, R. J. *et al.* Pathomic fusion: an integrated framework for fusing histopathology and genomic features for cancer diagnosis and prognosis. *IEEE Transactions on Medical Imaging* **41**, 757-770 (2020).

## Data availability

The research data involved in this study is all publicly available and can be downloaded from the following link: https://huggingface.co/datasets/MahmoodLab/hest.

## Acknowledgements

This research is partially funded by the China Scholarship Council (CSC).

## Author contributions

J.L. and D.M. initiated and designed the study. J.L. was responsible for writing the code, processing the data and performing the experiments. M.E. and F.F. provided guidance and insights from medical and clinical perspectives. Z.W. and D.M. reviewed the content related to artificial intelligence in the manuscript. M.E. and F.F. reviewed the content related to medicine and spatial transcriptomics in the manuscript. J.L. wrote the first draft of the manuscript. All authors contributed to the writing and editing of the revised manuscript and approved the manuscript.

## Competing interests

All authors declare no financial or non-financial competing interests.